\documentclass[runningheads]{llncs}
 \usepackage{booktabs} 
\usepackage[T1]{fontenc}
\usepackage{amssymb}
\usepackage{amsmath}
\usepackage{graphicx,verbatim}
\usepackage{xcolor}
\usepackage{pifont}
\newcommand{\cmark}{{\color{green!60!black}\ding{51}}}
\newcommand{\xmark}{{\color{red!70!black}\ding{55}}}
\begin{document}
\title{Geometry OR Tracker: Universal Geometric Operating Room Tracking}

\begin{comment}   Removed for anonymized MICCAI 2025 submission
\author{Yihua Shao\inst{1}\orcidID{0000-1111-2222-3333} \and
Kang Chen\inst{2,3}\orcidID{1111-2222-3333-4444} \and
Feng xue\inst{3}\orcidID{2222--3333-4444-5555} \and
Siyu Chen\inst{3}\orcidID{2222--3333-4444-5555}\and
Jinlin Wu\inst{3}\orcidID{2222--3333-4444-5555}\and
Zhen Chen\inst{3}\orcidID{2222--3333-4444-5555}\and
Zhen Lei\inst{3}\orcidID{2222--3333-4444-5555}}

\authorrunning{F. Author et al.}
 First names are abbreviated in the running head.

\institute{Princeton University, Princeton NJ 08544, USA \and
Springer Heidelberg, Tiergartenstr. 17, 69121 Heidelberg, Germany
\email{lncs@springer.com}\\
\url{http://www.springer.com/gp/computer-science/lncs} \and
ABC Institute, Rupert-Karls-University Heidelberg, Heidelberg, Germany\\
\email{\{abc,lncs\}@uni-heidelberg.de}}

\author{Yihua Shao\inst{1}\orcidID{0000-1111-2222-3333} \and
Kang Chen\inst{2,3}\orcidID{1111-2222-3333-4444} \and
Feng xue\inst{3}\orcidID{2222--3333-4444-5555} \and
Siyu Chen\inst{3}\orcidID{2222--3333-4444-5555}\and
Jinlin Wu\inst{3}\orcidID{2222--3333-4444-5555}\and
Zhen Chen\inst{3}\orcidID{2222--3333-4444-5555}\and
Zhen Lei\inst{3}\orcidID{2222--3333-4444-5555}}
\end{comment}

% \author{Anonymized Authors}  %% Added for anonymized MICCAI 2025 submission
% \authorrunning{Anonymized Author et al.}
% \institute{Anonymized Affiliations \\
%     \email{email@anonymized.com}}
\author{Yihua Shao\inst{1,2} \and
Kang Chen\inst{3} \and
Feng xue\inst{4} \and
Siyu Chen\inst{2}\and
Long Bai\inst{5}\and
Hongyuan Yu\inst{6}\and
Hao Tang\inst{3}\and
Jinlin Wu\inst{1,2}\and
Nassir Navab\inst{7}}
\institute{Centre for Artificial Intelligence and Robotics
Hong Kong Institute of Science and Innovation Chinese Academy of Sciences\and
State Key Laboratory of Multimodal Artificial Intelligence Systems (MAIS)
Institute of Automation
Chinese Academy of Sciences (CASIA)\and
Peking University\and
University of Trento, Trento, Italy\and
Alibaba DAMO Academy\and
Xiaomi Inc.\and
Technische Universität München\\
\email{feng.xue@unitn.it, jinlin.wu@cair-cas.org.hk}}
\maketitle              % typeset the header of the contribution
\begin{abstract}
In operating rooms (OR), world-scale multi-view 3D tracking supports downstream applications such as surgeon behavior recognition, where physically meaningful quantities such as distances and motion statistics must be measured in meters.
However, real clinical deployments rarely satisfy the geometric prerequisites for stable multi-view fusion and tracking: camera calibration and RGB--D registration are always unreliable, leading to cross-view geometric inconsistency that produces ``ghosting'' during fusion and degrades 3D trajectories in a shared OR coordinate frame.
To address this, we introduce \textbf{Geometry OR Tracker}, a two-stage pipeline that first rectifies imprecise calibration into a scale-consistent and geometrically consistent camera setup with a single global scale via a \emph{Multi-view Metric Geometry Rectification} module, and then performs \emph{Occlusion-Robust 3D Point Tracking} directly in the unified OR world frame.
On the MM-OR benchmark, improved geometric consistency translates into tracking gains: our rectification front-end reduces cross-view depth disagreement by more than $30\times$ compared to raw calibration.
Ablation studies further demonstrate the relationship between calibration quality and tracking accuracy, showing that improved geometric consistency yields stronger world-frame tracking.

\keywords{Operating Room \and Multi-view 3D Tracking \and Geometry Rectification.}
\end{abstract}
\section{Introduction}
Tracking dynamic objects in operating rooms (OR) is a prerequisite for emerging clinical applications, including VR-assisted surgery, automated workflow analysis, and surgeon behavior recognition~\cite{luiten2024dynamic,wang2025shape,seidenschwarz2025dynomo}.
Due to the confined space and frequent mutual occlusions inherent to OR~\cite{chadebecq2020computer,eck2023real,gerats2023dynamic}, multi-view camera arrays provide complementary viewpoints that can resolve occlusions and enable metric 3D trajectories.
A key missing capability is metric and time-resolved 3D understanding: to measure distances, velocities, and interactions reliably for dynamic OR entities, we need a consistent metric 4D scene representation and stable 3D trajectories in a shared OR coordinate frame.

Recent point-level 2D trackers~\cite{doersch2022tap,karaev2025cotracker3} provide strong image-space correspondences but cannot achieve metric consistency across views. This motivates 3D tracking, where geometric constraints can fuse multi-view evidence and provide metrically meaningful motion. However, monocular 3D trackers~\cite{xiao2024spatialtracker,xiao2025spatialtrackerv2,ngo2024delta} remain vulnerable to scale ambiguity and drift, and multi-view methods~\cite{wang2025scenetracker,rajic2025mvtracker} rely on accurate intrinsics, and aligned RGB-D. In practice, OR calibration is hard to obtain and maintain due to placement error and temporal drift~\cite{fluckiger2025autocalib,obayashi2023surgicalcalib}. RGB-D misalignment and partial coverage are common, so even moderate errors can induce fusion artifacts that destabilize tracking~\cite{nwoye2023cholectrack20,fu2020joint}. Moreover, in ORs, artifacts amplify calibration noise and break long-horizon metric consistency. Geometry foundation models~\cite{wu2024teleor,keetha2025mapanything,wang2025vggt,mildenhall2020nerf,gerats2025nerf} suggest a path toward stronger geometric priors, but their use for robust tracking under noisy OR calibrations remains unexplored, motivating a method that enforces universal geometric consistency while tolerating imperfect calibration.

In this work, we investigate the role of geometric consistency in operating-room tracking and show that robust metric 3D tracking is often bottlenecked by calibration quality rather than correspondence modeling alone.
We present \textbf{Geometry OR Tracker}, a two-stage pipeline that explicitly turns noisy OR calibration into \emph{tracking-ready} metric geometry and then performs occlusion-robust world-frame tracking, as illustrated in Fig.~\ref{fig:pipeline}.
In Stage 1, a \emph{Multi-view Metric Geometry Rectification} module treats the provided intrinsics/extrinsics and RGB--D alignment as unpresice priors and predicts a self-consistent camera bundle together with globally scaled metric depth, substantially reducing cross-view misregistration.
In Stage 2, we perform \emph{Occlusion-Robust Metric 3D Point Tracking} by lifting multi-view observations into a fused 3D feature cloud in the rectified OR frame and updating point trajectories via local 3D neighborhood reasoning and iterative refinement.

On the MM-OR benchmark, this geometry-to-tracking design yields consistent gains over representative baselines across standard 3D tracking metrics, while also improving cross-view geometric agreement during multi-view fusion.
Ablations further quantify how rectification inputs affect both metric depth quality and downstream tracking, revealing that improving geometric consistency is strongly correlated with better world-frame trajectories.

Our contributions are as follows:
\begin{itemize}
    \item \textbf{A calibration-robust pipeline} for multi-view metric 3D tracking in ORs that produces tracking-ready geometry from noisy real-world calibration and RGB--D misregistration.
    \item \textbf{A geometry--tracking study} that empirically demonstrates a strong correlation between geometric consistency and downstream tracking accuracy, and identifies which rectification cues matter most in practice.
    \item \textbf{Strong performance on MM-OR} with consistent improvements over competitive multi-view and single-view tracking baselines in multiple evaluation metrics.
\end{itemize}

\section{Methodology}

\noindent{\textbf{Overview}}
Fig.~\ref{fig:pipeline} shows our two-stage framework:
Stage~1, \emph{Multi-view Metric Calibration Rectification}, refines calibration, estimates per-frame depth, and recovers a global metric scale.
Stage~2, \emph{Occlusion-Robust Metric 3D Point Tracking}, fuses rectified views into a 3D feature cloud and tracks 3D query points in the OR frame via local 3D retrieval and iterative refinement for dynamic entities such as instruments and staff.
\subsection{Problem Formulation in the OR Context}
\label{sec:problem_setup}
We consider an operating room equipped with $V$ synchronized RGB--D cameras. In time $t$ and view $v$, the system captures an RGB frame $I_t^v$ and a depth map $D_t^v$.
We track $N$ clinically meaningful 3D targets, such as tool tips or keypoints on surgical staff. Each query
$\mathbf{q}^n=[t_q^n, x_q^n, y_q^n, z_q^n]^\top$
specifies the first observation time and a 3D location in a shared room coordinate frame. The goal is to estimate the metric 3D trajectory
$\mathcal{P}^n=\{\mathbf{p}_t^n\in\mathbb{R}^3\}_{t\ge t_q^n}$
and visibility
$\mathcal{V}^n=\{v_t^n\in[0,1]\}_{t\ge t_q^n}$,
where a larger $v_t^n$ indicates a higher confidence that the point is visible in at least one view at time $t$.

Each camera has intrinsic $K^v$ and extrinsic $P^v=[R^v|T^v]$. For a pixel $\mathbf{u}=[u_x,\,u_y,\,1]^\top$ with depth $d$, the camera-frame point is
$\mathbf{x}=d\,(K^v)^{-1}\mathbf{u}$,
and the corresponding room-frame point is
$\mathbf{X}=R^v\mathbf{x}+T^v$.

\begin{figure}[t]
    \centering
    \includegraphics[width=\linewidth]{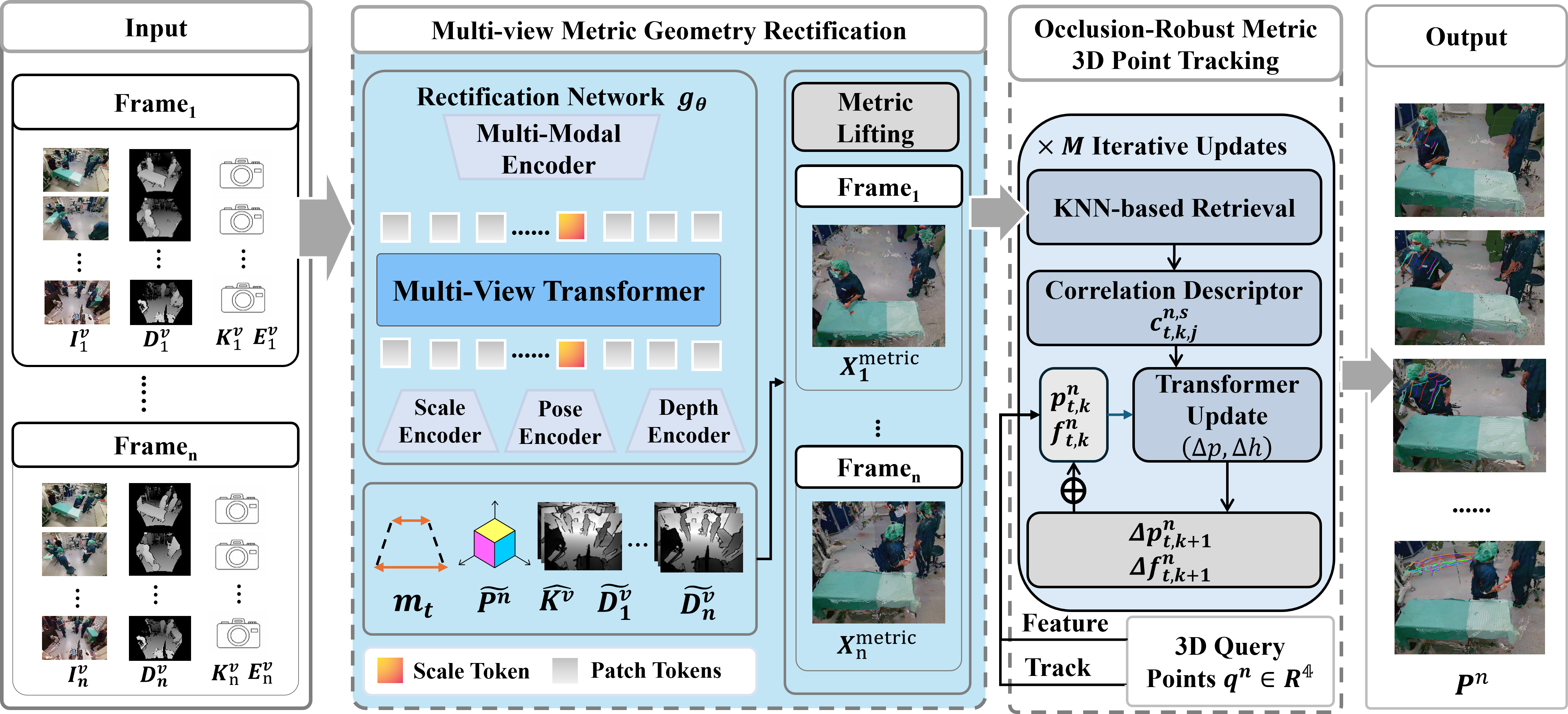}
    \caption{Pipeline of \textbf{Geometry OR Tracker}. Our method decouples geometry rectification from downstream tracking. \textbf{Multi-view Metric Calibration Rectification} produces a metric-consistent camera calibration and per-frame depth with a global metric scale; \textbf{Occlusion-Robust Metric 3D Point Tracking} then lifts multi-view observations into a fused 3D feature point cloud and tracks 3D query points in a shared room coordinate frame via local 3D neighborhood retrieval and iterative refinement.}
    \label{fig:pipeline}
    \vspace{-0.5cm}
\end{figure}

In practice, OR data suffer from frequent occlusions and calibration noise due to constrained camera placement and drift. Consequently, the provided $K^v$ and $P^v$ are often inaccurate, and lifting RGB--D observations directly into 3D yields an inconsistent metric space with ghosting and misaligned point clouds, which destabilize correspondence and trajectory estimation.

As shown in Fig.~\ref{fig:pipeline}, our method rectifies calibration and depth, then performs tracking in the rectified metric space. The rectification stage takes the imprecise calibration and depth as input, and refines them into a metrically consistent camera geometry and per-frame depth in OR frame. The tracking stage lifts multi-view observations into a fused 3D feature cloud and updates 3D correspondences to remain robust to occlusions and view-dependent changes.

\subsection{Multi-view Metric Geometry Rectification}
\label{sec:camera_rectification}
Since OR cameras are typically static within a procedure, we estimate a single globally rectified calibration from the first synchronized multi-view frame and apply it to the entire sequence. Otherwise, per-frame reconstruction would yield slightly different predicted intrinsics and extrinsics across frames, which may lead to temporal inconsistency in camera geometry and unnecessary drift in downstream reconstruction and tracking.
We used a geometry foundation model~\cite{keetha2025mapanything} as a priori to mitigate OR-specific calibration noise and to output tracking-ready metric geometry.
Imprecise calibration inputs, such as intrinsic, extrinsic, and depth, are optional, but they have a substantial impact on reconstruction quality.
Unless otherwise stated, we use RGB images together with intrinsics and depth as rectification input.

\noindent\textbf{Rectified metric outputs.}
Given the RGB set with multiple views for the first-frame $\mathcal{I}_1=\{I_1^v\}_{v=1}^{V}$ and optional geometric hints $\{K_1^v, P_1^v, D_1^v\}$ that may be available for a subset of views, the rectifier predicts a global metric scale $m$, rectified intrinsics $\tilde{\mathbf{K}}^{v}$, rectified camera-to-room poses $\tilde{\mathbf{P}}^{v}$, and rectified depth maps per-frame $\tilde{\mathbf{D}}_t^{v}$:
\begin{equation}
g_{\theta}\!\Bigl(\mathcal{I}_1,\ [\mathcal{K}_1,\mathcal{E}_1,\mathcal{D}_1]\Bigr)
=
\Bigl(m,\ \{\tilde{\mathbf{K}}^{v},\tilde{\mathbf{P}}^{v}\}_{v=1}^{V},\ \{\tilde{\mathbf{D}}_t^{v}\}_{t=1}^{T}\Bigr).
\label{eq:factored_pred_static}
\end{equation}
The predicted global metric-scale $m$ ties the reconstruction to real-world metric units, enabling geometrically faithful point maps for downstream fusion and tracking.

\noindent\textbf{Metric lifting and point maps.}
With rectified calibration and metric scale, each pixel corresponds to a physically meaningful geometry.
For any frame $t$ and view $v$, let $\mathbf{u}=[u_x,u_y,1]^\top$ denote a homogeneous pixel coordinate and let $\tilde{D}_t^v(\mathbf{u})$ denote its depth.
We first back-project to a 3D point in the camera frame:
\begin{equation}
\tilde{\mathbf{l}}_t^{v}(\mathbf{u})
=
\tilde{D}_t^{v}(\mathbf{u})\ (\tilde{\mathbf{K}}^{v})^{-1}\mathbf{u}
\in\mathbb{R}^{3}.
\label{eq:local_pointmap}
\end{equation}
We then transform it into the shared room coordinate frame using the rectified pose
$\tilde{\mathbf{P}}^{v}=\begin{bmatrix}\tilde{\mathbf{R}}^{v} & \tilde{\mathbf{t}}^{v}\\ \mathbf{0}^{\top}&1\end{bmatrix}$:
\begin{equation}
\tilde{\mathbf{x}}_t^{v}(\mathbf{u})
=
\tilde{\mathbf{R}}^{v}\,\tilde{\mathbf{l}}_t^{v}(\mathbf{u})
+
\tilde{\mathbf{t}}^{v}
\in \mathbb{R}^{3}.
\label{eq:world_pointmap}
\end{equation}
Finally, we lift to metric scale with the global factor $m$:
\begin{equation}
\mathbf{x}_{t}^{\mathrm{metric},v}(\mathbf{u})
=
m\,\tilde{\mathbf{x}}_t^{v}(\mathbf{u}),
\qquad
X_t=\bigcup_{v=1}^{V}\{\mathbf{x}_{t}^{\mathrm{metric},v}(\mathbf{u})\}.
\label{eq:metric_lift}
\end{equation}
These per-frame metric point maps $X_t$ provide a stable geometric interface for the tracking stage, and substantially reduce cross-view misalignment that would otherwise corrupt correspondence under occlusion.

\subsection{Occlusion-Robust Metric 3D Point Tracking}
\label{sec:tracking}
With clean metric room frames established, we perform metric 3D point tracking conditioned on the rectified geometry using a pretrained multi-view  tracker~\cite{rajic2025mvtracker}.
The key clinical benefit is continuity under occlusions.
When a target is blocked in some views by staff or instruments, the tracker can still update the trajectory using evidence from other views that project to the same 3D region.

\noindent\textbf{Lift multi-view evidence into a fused 3D feature cloud.}
For each view $v$ at time $t$, a backbone extracts 2D feature maps denoted $\mathbf{F}^{v}_t$.
Using the rectified metric geometry in Eq.~\eqref{eq:metric_lift}, each valid pixel corresponds to a metric 3D point and a feature vector sampled in that pixel.
Aggregating all views yields a fused 3D feature point cloud:
\begin{equation}
\mathcal{Y}_t=\Bigl\{\bigl(\mathbf{y}_{t,j},\ \mathbf{f}_{t,j}\bigr)\Bigr\}_{j=1}^{M_t},\qquad
\mathbf{y}_{t,j}\in\mathbb{R}^{3},\ \mathbf{f}_{t,j}\in\mathbb{R}^{C}.
\end{equation}

\noindent\textbf{Local 3D neighborhood retrieval for correspondence.}
Given a current estimate $\mathbf{p}^{n}_{t}$ for track $n$, we retrieve a local neighborhood in a 3D metric space:
\begin{equation}
\mathcal{N}_K(\mathbf{p}^{n}_{t})
=
\mathrm{kNN}\!\left(\mathbf{p}^{n}_{t},\ \{\mathbf{y}_{t,j}\}_{j=1}^{M_t}\right).
\end{equation}
This neighborhood remains geometrically meaningful when the dominant view changes, because all cameras contribute points to the same room coordinate frame.

\noindent\textbf{Iterative refinement and visibility.}
We update track positions over time with an iterative transformer refinement module following prior point trackers~\cite{karaev2024cotrackerbettertrack,rajic2025mvtracker}.
Given fused 3D clouds and the query initialization $\mathbf{q}^n$, the tracker outputs the trajectory and visibility:
\begin{equation}
\{\mathbf{p}_t^n, v_t^n\}_{t=t_q^n}^{T}
=
h_{\phi}\!\left(\{\mathcal{Y}_t\}_{t=t_q^n}^{T},\ \mathbf{q}^n\right).
\end{equation}

\section{Experiments}

\subsection{Experimental Setup}

\textbf{Dataset.} We evaluate our method on the MM-OR dataset~\cite{ozsoy2025mm}, which provides fully synchronized multi-view RGB--D videos with camera calibration. We used the data frome five Kinect cameras from the multi-view RGB--D subset and conducted all experiments on 10 randomly selected scenes, each with 100 frames.

\noindent\textbf{Baselines.}
For \emph{Multi-view Metric Calibration Rectification(MMCR)}, we use dataset provided raw calibration and depth as the baseline.
For \emph{3D point tracking} using robust obstruction metrics, we compare with representative point trackers, including CoTracker3~\cite{karaev2025cotracker3}, LocoTrack~\cite{cho2024local}, SpaTrackerV2~\cite{xiao2025spatialtrackerv2}, SceneTracker~\cite{wang2025scenetracker}, DELTA~\cite{ngo2024delta}, and MVTracker~\cite{rajic2025mvtracker}.
All baselines except MVTracker are applied per view due to their single-view design. We report their per-view 2D tracking results to provide a fair comparison against multi-view 3D baselines under the same input setting.

\noindent\textbf{Evaluation Metrics.}
We report (i) \textbf{point tracking} quality in the shared OR frame and (ii) \textbf{metric scale recovery} for the rectified geometry.
For tracking, we use Average Jaccard (AJ), threshold-averaged accuracy ($\Delta_{\text{avg}}$), Occlusion Accuracy (OA), and Median Trajectory Error (MTE).
For scale recovery, we compare the rectified depth with the ground truth using the AbsRel and RMSE.

\noindent\textbf{Implementation Details.}
We use the rectifier's first-frame prediction as a sequence-level calibration and keep it fixed for the whole sequence.
After rectifying calibration and depth with MapAnything~\cite{keetha2025mapanything}, we fuse multi-view RGB-D observations into a metric 3D point cloud with features and run MVTracker~\cite{rajic2025mvtracker} as our tracking backend to obtain metric 3D trajectories in the shared OR frame.
All experiments were conducted on an NVIDIA GeForce RTX 4090 GPU.

\subsection{Main Results}

\begin{table}[t]
\centering
\caption{\textbf{Geometry Consistency via Cross-view Depth Reprojection.}}
\begin{tabular}{lcc}
\toprule
\textbf{Method} & \textbf{Mean Error (m)$\downarrow$} & \textbf{Median Error (m)$\downarrow$} \\
\midrule
Raw Calibration & 1.4115 & 1.4151 \\
Our Method      & 0.0459 & 0.0204 \\
\bottomrule
\end{tabular}
\label{tab:depth_reproj_consistency}
\end{table}

\noindent \textbf{Metric Calibration Rectification Results.}
Table~\ref{tab:depth_reproj_consistency} reports cross-view depth reprojection consistency. Compared to raw calibration, our rectification substantially reduces depth disagreement, lowering both mean and median errors. These results indicate markedly improved geometric alignment across views.
We attribute this gain to the rectification stage explicitly enforcing multi-view metric consistency by refining intrinsics and poses while correcting RGB–D misalignment and scale drift. Consequently, corresponding surfaces are reconstructed in a coherent room frame, yielding significantly smaller cross-view depth discrepancies, more stable fusion, and a marked reduction of ghosting artifacts.

\begin{figure}[t]
    \centering
    \includegraphics[width=\linewidth]{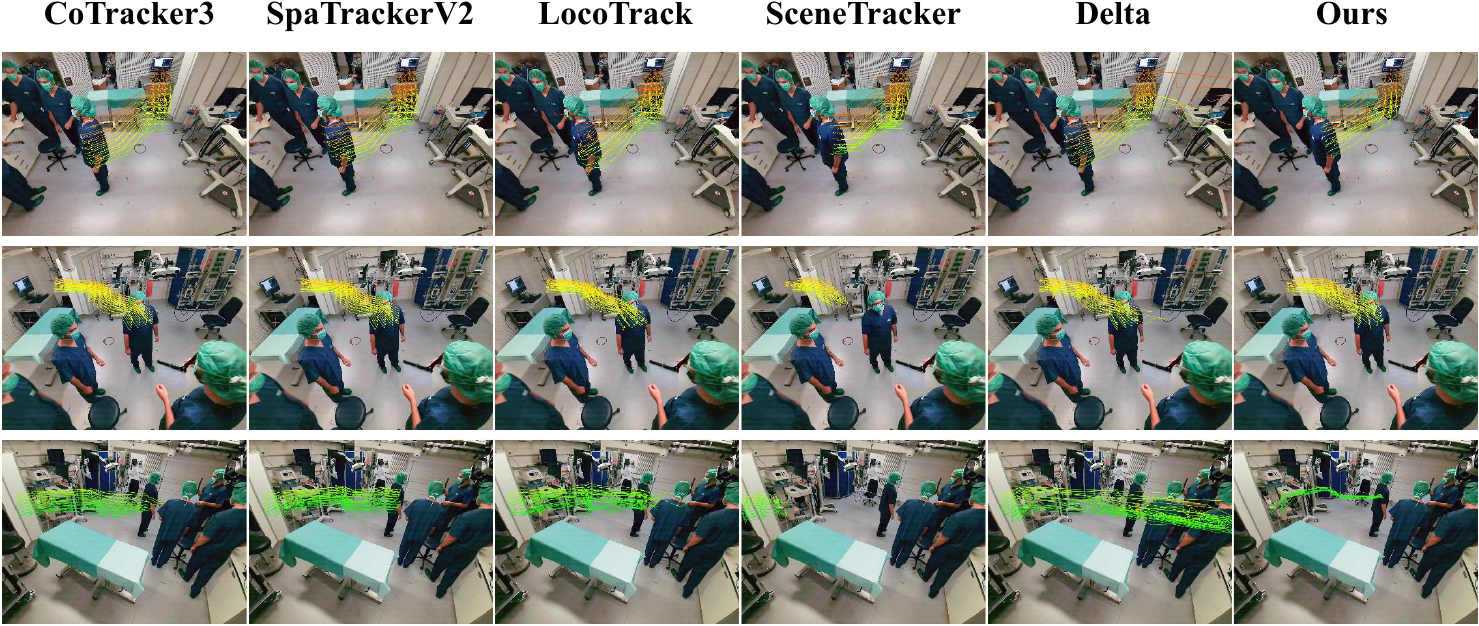}
    \caption{\textbf{2D qualitative comparison of tracking} between baselines and our method.}
    \label{fig:2D}
    \vspace{-0.3cm}
\end{figure}

\noindent \textbf{Metric 3D Point Tracking Results.}
Table~\ref{tab:mmor_tracking} reports quantitative 3D tracking performance in MM-OR using AJ, $\Delta_{\text{avg}}$, OA, and MTE.
Among all metrics, our method achieves the best AJ and $\Delta_{\text{avg}}$, indicating more accurate 3D localization over time and across tolerance thresholds.
We attribute these gains to multi-view metric 3D tracking that leverages cross-view redundancy under occlusions, and to geometry rectification that enforces cross-view consistency and suppresses fusion ``ghosting'' from unprecise calibration or RGB--D misalignment.
Moreover, we obtain the best OA, suggesting improved robustness under occlusions, and the lowest MTE, reflecting reduced trajectory drift in the shared OR frame.
Furthermore, Fig.~\ref{fig:2D} also shows that our method delivers the best overall tracking quality: trajectories remain tight and temporally stable on the target even under heavy occlusions, while 2D per-view baselines often drift, or break due to view-specific visibility changes and geometric inconsistency.

\begin{table*}[t]
\caption{\textbf{Tracking Results on MM-OR.} Best results are \textbf{bolded}, second best are \underline{underlined}.}
\label{tab:mmor_tracking}
\vskip 0.12in
\centering
\setlength{\tabcolsep}{7pt}
\renewcommand{\arraystretch}{1.05}
\begin{tabular}{lcccc}
\toprule
\textbf{Method} & \textbf{AJ $\uparrow$} & \textbf{$\Delta_{\text{avg}}$ $\uparrow$} & \textbf{OA $\uparrow$} & \textbf{MTE $\downarrow$} \\
\midrule
CoTracker3~\cite{karaev2025cotracker3}         & 79.67 & 85.16 & 83.42 & 7.13 \\
SpaTrackerV2~\cite{xiao2025spatialtrackerv2}   & 81.31 & 83.77 & 89.82 & 9.25 \\
LocoTrack~\cite{cho2024local}                  & 81.58 & 89.60 & 79.70 & 8.68 \\
SceneTracker~\cite{wang2025scenetracker}       & 69.03 & 86.47 & 73.75 & 6.80 \\
DELTA~\cite{ngo2024delta}                      & 77.88 & 87.29 & 73.18 & 9.46 \\
MVTracker~\cite{rajic2025mvtracker}            & \underline{84.78} & \underline{90.29} & \underline{93.18} & \underline{3.70} \\
\midrule
\textbf{Ours (Full)}                           & \textbf{89.73} & \textbf{93.65} & \textbf{96.28} & \textbf{3.46} \\
\midrule
w/o. MMCR (raw geometry)                        & \underline{84.78} & \underline{90.29} & \underline{93.18} & \underline{3.70} \\
\bottomrule
\end{tabular}
%\vspace{-0.5cm}
\end{table*}

\subsection{Ablation Study}

\noindent\textbf{Effect of Rectification.}
To quantify the impact of \emph{Multi-view Metric Calibration Rectification (MMCR)}, we remove it and run our tracker on geometry derived from the dataset raw geometry.
As shown in Table~\ref{tab:mmor_tracking}, disabling MMCR degrades tracking improves AJ, $\Delta_{\text{avg}}$, and OA while reducing MTE.
We attribute this to cross-view inconsistency and fusion ``ghosting'' induced by raw calibration/RGB--D misalignment. MMCR rectifies the geometry into a coherent metric setup, yielding cleaner 3D evidence for correspondence estimation in the shared OR frame.

\noindent\textbf{Scale Recovery under Different Inputs.}
To identify the most effective inputs for rectification, we evaluate scale recovery under different input combinations by comparing predicted depth maps to GT depth in MM-OR.
As shown in Table~\ref{tab:rectif_depth_tracking_ablation}, using RGB and depth is critical, and adding intrinsics further improves AbsRel/RMSE by reducing scale ambiguity and depth distortion through stronger geometric constraints.

\noindent\textbf{Downstream 3D Point Tracking.}
We further run \emph{Occlusion-Robust Metric 3D Point Tracking} on the geometry produced by different rectification inputs.
As shown in Table~\ref{tab:rectif_depth_tracking_ablation}, better rectification yields better tracking, since improved cross-view metric consistency reduces ``ghosting'' and stabilizes local 3D neighborhood retrieval and refinement under occlusions.
Notably, configurations that achieve lower depth errors(AbsRel/RMSE) also tend to produce higher AJ/$\Delta_{\text{avg}}$/OA and lower MTE, indicating a strong coupling between metric geometric fidelity and tracking robustness in the shared OR frame.
\makeatletter
\@ifpackageloaded{xcolor}{}{\usepackage{xcolor}}
\@ifpackageloaded{pifont}{}{\usepackage{pifont}}
\@ifpackageloaded{booktabs}{}{\usepackage{booktabs}}
\makeatother

% safety: if xcolor isn't loaded for some reason, don't print color names
\providecommand{\textcolor}[2]{#2}

%\newcommand{\cmark}{\textcolor{green}{\ding{51}}} % ✓
%\newcommand{\xmark}{\textcolor{red}{\ding{55}}}   % ✗

% ============================
% 2) TABLE 3
% ============================
\begin{table*}[t]
\caption{\textbf{Effects of Different Input.}
AbsRel/RMSE evaluates depth accuracy; AJ/$\Delta_{\text{avg}}$/OA/MTE evaluate downstream 3D point tracking.
}
\label{tab:rectif_depth_tracking_ablation}
\vskip 0.10in
\centering
\resizebox{\textwidth}{!}{%
\begin{small}
\setlength{\tabcolsep}{5.2pt}
\renewcommand{\arraystretch}{1.10}
\begin{tabular}{@{}cccccccccc@{}}
\toprule
\multicolumn{4}{c}{\textbf{Rectification Inputs}} & \multicolumn{2}{c}{\textbf{Depth}} & \multicolumn{4}{c}{\textbf{Tracking}} \\
\cmidrule(lr){1-4}\cmidrule(lr){5-6}\cmidrule(lr){7-10}
\textbf{RGB} & \textbf{K} & \textbf{Pose} & \textbf{Depth} &
\textbf{AbsRel $\downarrow$} & \textbf{RMSE $\downarrow$} &
\textbf{AJ $\uparrow$} & \textbf{$\Delta_{\text{avg}}$ $\uparrow$} & \textbf{OA $\uparrow$} & \textbf{MTE $\downarrow$} \\
\midrule
\cmark & \xmark & \xmark & \xmark & 0.107 & 0.710 & 41.95 & 64.18 & 64.88 & 11.44 \\
\cmark & \cmark & \xmark & \xmark & 0.143 & 0.739 & 44.47 & 67.47 & 65.23 & 10.50 \\
\cmark & \xmark & \cmark & \xmark & 0.144 & 0.755 & 39.77 & 69.44 & 59.98 & 9.23 \\
\cmark & \cmark & \cmark & \xmark & 0.143 & 0.770 & 40.37 & 72.85 & 58.84 & 8.01 \\
\cmark & \cmark & \cmark & \cmark & \underline{0.060} & \underline{0.653} & \underline{74.78} & \underline{85.18} & \underline{87.32} & \underline{6.11} \\
\cmark & \cmark & \xmark & \cmark & \textbf{0.057} & \textbf{0.625} & \textbf{89.73} & \textbf{93.65} & \textbf{96.28} & \textbf{3.46} \\
\bottomrule
\end{tabular}
\end{small}
\!}%
\vspace{-0.5cm}
\end{table*}

\section{Conclusion}
In this paper, we present \textbf{Geometry OR Tracker}, a geometry-consistent framework for calibration-robust multi-view RGB-D 4D reconstruction and 3D point tracking. By introducing \emph{Multi-view Metric Calibration Rectification}, which enforces multi-view geometric consistency, our method corrects unperspicuous intrinsics and extrinsics and provides tracking-ready metric geometry for \emph{Occlusion-Robust Metric 3D Point Tracking}. In experiments, we show that \emph{Geometry-Rectified OR Tracker} achieves strong 3D point tracking performance and consistently outperforms representative baselines across standard metrics. In summary, we solve calibration-induced geometric inconsistency in multi-view RGB-D operating rooms, enabling reliable metric 4D reconstruction and 3D point tracking under noisy and unpresice real-world calibrations.

\bibliographystyle{splncs04}
\bibliography{mybibliography}

\end{document}